\pdfoutput=1
\PassOptionsToPackage{table}{xcolor}

\documentclass[11pt]{article}

\usepackage{acl}

\usepackage{times}
\usepackage{latexsym}
\usepackage{longtable}
\usepackage{booktabs}
\usepackage{hyperref}
\definecolor{cogsetA}{gray}{0.9}  % light gray
\definecolor{cogsetB}{gray}{0.8}  % slightly darker gray
\usepackage{siunitx}
\usepackage{graphicx}
\graphicspath{{img/}}
 \usepackage{microtype}

\usepackage[english]{babel} % English as the main language
\usepackage{newtxtext,newtxmath}
\usepackage{tipa}

\title{Beyond cognacy\thanks{This is an updated version of the paper published in the Proceedings of SIGTYP 2025. After publication, three defects were found in the implementation of the MSA method: a sign error in the null-model training objective, two bugs in the T-Coffee implementation, and a flawed validation objective in the pHMM training, which had caused the published pHMM to degenerate into a measure of symbol identity. For the present version, the pHMM was retrained, the T-Coffee guide tree was made independent of the methods under comparison, and the entire pipeline was re-run on a current version of Lexibank, for all three methods, with a fixed random seed. The dataset thereby grew from 928 to 974 languages and from 14 to 15 families. All reported numbers changed; in addition, the difficulty scores are now computed with the Pythia predictor built into RAxML-NG and are not comparable to the previously reported ones. The qualitative conclusions are unchanged.}}

\author{Gerhard Jäger\\
  University of Tübingen, Germany\\
  \texttt{gerhard.jaeger@uni-tuebingen.de}\\
    \emph{August 7, 2026}\thanks{Published in \emph{Proceedings of The 7th Workshop on Research in Computational Linguistic Typology and Multilingual NLP}}
  }

\begin{document}

\maketitle
\begin{abstract}
    Computational phylogenetics has become an established tool in historical linguistics, with many language families now analyzed using likelihood-based inference. However, standard approaches rely on expert-annotated cognate sets, which are sparse, labor-intensive to produce, and limited to individual language families. This paper explores alternatives by comparing the established method to two fully automated methods that extract phylogenetic signal directly from lexical data. One uses automatic cognate clustering with unigram/concept features; the other applies multiple sequence alignment (MSA) derived from a pair-hidden Markov model. Both are evaluated against expert classifications from Glottolog and typological data from Grambank. Also, the intrinsic strengths of the phylogenetic signal in the characters are compared. Results show that MSA-based inference yields trees more consistent with linguistic classifications, better predicts typological variation, and provides a clearer phylogenetic signal, suggesting it as a promising, scalable alternative to traditional cognate-based methods. This opens new avenues for global-scale language phylogenies beyond expert annotation bottlenecks.
\end{abstract}

  \section{Introduction}

Originally developed in computational biology, quantitative methods for phylogenetic reconstruction using likelihood-based inference frameworks have now gained widespread acceptance in comparative linguistics. This is evident from the growing number of computational phylogenies proposed for some of the world's largest language families, including Dravidian \citep{Kolipakam2018}, Sino-Tibetan \citep{Sagart2019}, and Indo-European \citep{Heggarty2023}. Moreover, fully automated approaches — where even cognate identification is performed algorithmically — have demonstrated a surprising degree of robustness \citep{rama2018automatic}. In contrast to the pre-computational era of historical linguistics, where such detailed reconstructions were rare, the generation of fully resolved phylogenies with branch lengths and, in some cases, estimated divergence dates has now become a standard practice in studies of language evolution.

Despite the increasing recognition of computational language phylogenies as a useful addition to the comparative linguistics toolkit, skepticism remains prevalent. A key concern raised by critics is that phylogenetic analyses are often based on cognate sets—groups of historically related words—extracted from semantically aligned word lists. Since these cognate sets are based on expert annotations, they are sparse, labor-intensive to acquire, and raise concerns regarding replicability.

Another limitation of phylogenetic inference based on cognate classes is that it is by definition constrained to individual language families. There is legitimate interest in automatically inferred trees spanning larger collections of languages, perhaps from the entire world. Such trees provide information about the strength of evidence for putative macro-families \citep{jager2015support,akavarapu-bhattacharya-2024-likelihood}. Furthermore, they are useful for downstream tasks such as the statistical modeling of global language evolution \citep{bentz2018evolution,bouckaert2022global}.

The literature contains several proposed workflows for extracting character matrices from word lists without cognate annotations, which can then used as input for likelihood-based phylogenetic inference. This paper presents a comparison of cognate-based phylogenetic inference with two such proposals, the one by \citet{jager2018global} and the one by \citet{akavarapu-bhattacharya-2024-likelihood}. These methods are evaluated in three ways: (1) by comparing the inferred phylogenies with the Glottolog expert classification \citep{hammarstrom2024glottolog}, (2) how well the inferred phylogenies fit to the typological features from \citep{grambank_release}, and (3) an estimation of the strength of the phylogenetic signal in the data, using the difficulty measure implemented in \emph{Pythia} \citep{haag2022easy}.

\section{Materials and Methods}

\subsection{Materials}

Word lists were obtained from Lexibank\footnote{\texttt{\url{https://github.com/lexibank}}; \citealt{List2023}}. These datasets contain lexical entries, including the language they belong to, their meaning, form in IPA transcription, and often a manual cognate annotation. The datasets are curated by the Lexibank community and are available in a standardized format, which makes them suitable for computational analyses.

In a first step, all datasets provided by the Lexibank organisation were downloaded. The IPA transcriptions were converted to the ASJP sound-class alphabet using the python package \emph{lingpy} \citep{LingPy2024}. After discarding entries without a Concepticon gloss or without a convertible transcription, this amounts to 2,332,222 lexical entries from 7,225 languages (identified by glottocodes), drawn from 142 datasets. Since identical word forms are often recorded by more than one dataset, entries agreeing in glottocode, concept, and ASJP transcription were deduplicated. This leaves 1,793,324 entries. 

For the purpose of evaluation, typological features were obtained from Grambank\footnote{\texttt{\url{https://github.com/grambank/grambank}}}. This results in 355,097 binary entries from 2,467 languages and 195 typological features.

A subset of Lexibank data was selected according to the following criteria:

\begin{itemize}
    \item The meaning of the entry belongs to the 110 concepts with the largest coverage.
    \item The entry comes from a language--dataset combination for which manual cognate annotations (\texttt{Cognateset\_ID}) are available.
    \item The entry comes from a language with a Glottocode that is present in the Grambank data.
    \item The language has entries for at least 40 of the 110 concepts.
\end{itemize}

This leaves 106,028 entries from 974 languages.

Constraining the Grambank data to these 974 languages leaves 141,375 binary data points from all 195 features.

The gold standard tree was obtained from Glottolog.\footnote{\texttt{\url{https://zenodo.org/records/10804582/files/glottolog/glottolog-cldf-v5.0.zip}}}

\subsection{Methods}

The overall workflow consists of the following steps:

\begin{enumerate}
    \item Phylogenetic inference
    \begin{enumerate}
    \item Generate a binary character matrix from the Lexibank data.
    \item Infer a phylogenetic tree from this character matrix.
    \end{enumerate}
    \item Evaluation
    \begin{enumerate}
    \item Compare the inferred phylogenetic tree with the Glottolog expert classification.
    \item Compare the inferred phylogenetic tree with the Grambank typological features.
    \item Assess the strength of the phylogenetic signal in the data.
    \end{enumerate}
\end{enumerate}

Three different methods were used to generate a binary character matrix: (1) binarized expert-annotated cognate classes, (2) a combination of automatic cognate clustering and unigram/concept features as described in \citet{jager2018global}, and (3) a variant of the method developed by \citet{akavarapu-bhattacharya-2024-likelihood} using multiple sequence alignment.

\subsubsection{Expert-annotated cognate classes (cc)}

Here we use the method introduced by \citet{Ringe2002} and \citet{gray2003language}. Each cognate class is treated as a character. A language is coded as 1 if it has a cognate in the class, 0 if it has a different cognate class for the same concept, and missing if it has no cognate for the concept. This results in a matrix with 974 rows and 29,695 columns.

Since each cognate class is, by definition, confined to a single language family, this character matrix contains no signal beyond the family level.

In the tables and figures below, this method is referred to as cc (for cognate classes).

\subsubsection{Automatic cognate clustering and unigram/concept features (PMI)}

The workflow proposed by \citet{jager2018global} was replicated. This approach uses two types of characters.

\begin{itemize}
    \item Binarized cognate classes obtained via automatic cognate clustering. This involves (1) supervised training of a Support Vector Machine classifier which takes a pair of words and predicts the labels 1 (cognate) or 0 (non-cognate), using manual cognate classification for supervision, (2) creating a distance matrix for all entries for a given concept from the 110 concepts defined above, and (3) clustering the distance matrix using the \emph{label propagation} algorithm \citep{raghavan2007near}.
    \item Unigram/concept characters. For each combination of a concept $c$ and an ASJP sound class $s$, a language is coded as 1 if it has a word for concept $c$ that contains sound class $s$, missing if it has no word for concept $c$, and 0 otherwise.
\end{itemize}

This resulted in a matrix with 974 rows and 49,441 columns.

Since the \emph{pointwise mutual information} between sound classes plays an essential role in this workflow, the method is referred to as PMI.

\subsubsection{Multiple sequence alignment (MSA)}

The method by \citet{akavarapu-bhattacharya-2024-likelihood} was used as starting point, but the present approach differs in various aspects. The method is based on the following steps:

In a first step, pairwise distances between languages in the full lexibank dataset were computed using the Levenshtein distance on the ASJP transcriptions and aggregating according to the method described in \citep{jager2018global}. Language pairs with a distance below $0.7$ were considered as \emph{probably related}, using the same heuristics as \citet{jager2018global}. All word pairs from such a language pair sharing their meaning are treated as \emph{potential cognates}. A random sample of 16 million potential cognates was assigned the label 1. As negative examples (label 0), an equal number of word pairs from probably related language pairs were sampled without the requirement that their meanings match.

In a second step, a classifier was trained on these labeled data. The classifier consists of a \emph{pair-Hidden Markov Model} (pHMM) \citep{durbin1998biological} and a logistic-regression layer. The classifier was trained for one epoch using the Adam optimizer, and the parameter state with the smallest loss on a held-out validation set was selected. The training was performed on a deduplicated wordlist that was derived from Lexibank in the way described above but covers a larger set of concepts; it is shared with a related project on world-wide language phylogenies. The resulting parameters of the pHMM were used in the next step.

A pHMM defines a probability distribution over pairs of aligned sequences of sound classes. This involves (1) emission probabilities for all pairs of sound classes that are matched in the alignment, (2) emission probabilities for individual sound classes if they are aligned with a gap, and (3) transition probabilities between the hidden states \emph{match}, \emph{gap in string 1}, \emph{gap in string 2}, and \emph{final state}. 

It is instructive to inspect the emission probabilities in the trained model. Since the joint probability of a match is dominated by the overall frequencies of the two sound classes involved, association strength is better measured by the pointwise mutual information (PMI), i.e., the log-ratio between the match emission probability and the product of the corresponding marginal probabilities. In Table \ref{tab:0} the ten sound classes with the strongest association with /p/ are shown for illustration. This ranking is in good agreement with linguistic intuitions about potential sound correspondences.

\begin{table}[ht]
    \centering
    \begin{tabular}{cc}
        \toprule
        \textbf{Sound class} & \textbf{Log-odds (PMI)} \\
        \midrule
        p & 2.99 \\
        6 & 1.64 \\
        f & 1.49 \\
        b & 1.16 \\
        v & 0.74 \\
        h & 0.46 \\
        u & 0.09 \\
        L & 0.02 \\
        2 & $-0.22$ \\
        E & $-0.27$ \\
        \bottomrule
    \end{tabular}
    \caption{The ten sound classes with the strongest association with /p/ in the trained pHMM, measured as the log-odds of a match against chance co-occurrence (pointwise mutual information). In List's version of ASJP, numerals 1--6 encode tone contours (level, rising, falling, etc.) and 7 encodes the glottal stop (\textglotstop).}
    \label{tab:0}
\end{table}

\begin{table}[ht]
    \centering
    \begin{tabular}{cc}
        \toprule
        \textbf{Sound class }&\textbf{Log-odds} \\
        \midrule
        7 & 0.92 \\
        ! & 0.63 \\
        w & 0.54 \\
        h & 0.52 \\
        m & 0.49 \\
        n & 0.43 \\
        y & 0.42 \\
        6 & 0.36 \\
        b & 0.32 \\
        d & 0.32 \\
        \bottomrule
    \end{tabular}
    \caption{The ten sound classes with the highest propensity of being matched with a gap in the trained pHMM, measured as the log-ratio between the gap emission probability and the marginal match emission probability of the sound class. The symbol \texttt{!} denotes a click.}
    \label{tab:01}
\end{table}

A high value here is to be interpreted as a high propensity of instances of these sound classes to participate either in insertion or deletion.

The trained pHMM assigns a probability to each pair of aligned sequences. Via the forward algorithm, the probability of a pair of sequences is computed as the sum of the probabilities of all possible alignments between these sequences.

Following \citet{durbin1998biological}, a null-model was trained additionally that assigns individual probabilities to both sequences, disregarding the order of sound classes. The log-odds ratio of a pair of words of being generated by the pHMM vs.\ the null model can interpreted as a measure of the similarity of the two words.

To illustrate this, a collection of ten words were chosen at random from the dataset which all have an edit distance of 1 to the word \emph{baba}, and their log-odds ratios with respect to \emph{baba} were computed. The results are shown in Table \ref{tab:02}. 
\begin{table}[ht]
    \centering
    \begin{tabular}{lc}
        \toprule
        \textbf{word} & \textbf{log-odds} \\
        \midrule
        eaba  & -18.59 \\
        bawa  & -19.06 \\
        bIba  & -19.40 \\
        raba  & -19.86 \\
        maba  & -19.95 \\
        naba  & -20.03 \\
        babau & -20.68 \\
        babae & -20.78 \\
        zaba  & -20.99 \\
        xaba  & -22.04 \\
        \bottomrule
    \end{tabular}
    \caption{Ten randomly chosen words with an edit distance of 1 from \emph{baba}, alongside with the predicted log-odds to \emph{baba}.}
    \label{tab:02}
\end{table}

This ranking is broadly consistent with linguistic intuitions about potential cognacy: substitutions involving phonetically similar segments, such as in \emph{bawa} or \emph{bIba}, receive higher log-odds than substitutions involving dissimilar ones, such as in \emph{xaba} or \emph{zaba}.

In a third step, the trained pHMM was used in combination with the Viterbi algorithm to obtain pairwise sequence alignments for all synonymous word pairs from different languages within the smaller dataset of 974 languages and 110 concepts. 

In a fourth step, the pairwise sequence alignments were aggregated to a \emph{multiple sequence alignment} (MSA) using the \emph{T-Coffee} algorithm \citep{notredame2000t}. T-Coffee performs progressive alignment along a guide tree. To keep this guide tree independent of all methods under comparison, it was computed by applying the BIONJ algorithm \citep{gascuel1997bionj} to the aggregated Levenshtein distances described above, restricted to the 974 selected languages, and rooted with the minimal ancestor deviation method \citep{tria2017phylogenetic}.

Note that all reflexes of a given concept are aligned within a single MSA, regardless of cognacy. Such an MSA implicitly contains information both about cognacy and about sound correspondences.

An example (for a much smaller dataset) is shown in Table \ref{tab:1} for illustration. These are the reflexes of the concept \emph{louse} from the Tungusic languages in the dataset.\footnote{The data are taken from \url{https://zenodo.org/records/13163376}, which is based on \citep{Oskolskaya2021}.}

\begin{table*}[ht]
    \centering
    \begin{tabular}{llcccccccc}
        \toprule
        \textbf{Language} & \textbf{Cognateset\_ID} & \textbf{1} & \textbf{2} & \textbf{3} & \textbf{4} & \textbf{5} & \textbf{6} & \textbf{7} & \textbf{8} \\\midrule
        Even     & 16\_lousen-37 & \cellcolor{cogsetA}- & \cellcolor{cogsetA}k & \cellcolor{cogsetA}u & \cellcolor{cogsetA}m & \cellcolor{cogsetA}- & \cellcolor{cogsetA}k & \cellcolor{cogsetA}- & \cellcolor{cogsetA}e \\
        Kilen    & 16\_lousen-37 & \cellcolor{cogsetA}q & \cellcolor{cogsetA}h & \cellcolor{cogsetA}u & \cellcolor{cogsetA}m & \cellcolor{cogsetA}I & \cellcolor{cogsetA}k & \cellcolor{cogsetA}- & \cellcolor{cogsetA}I \\
        Negidal  & 16\_lousen-37 & \cellcolor{cogsetA}- & \cellcolor{cogsetA}k & \cellcolor{cogsetA}u & \cellcolor{cogsetA}m & \cellcolor{cogsetA}- & \cellcolor{cogsetA}k & \cellcolor{cogsetA}- & \cellcolor{cogsetA}I \\
        Oroch    & 16\_lousen-37 & \cellcolor{cogsetA}- & \cellcolor{cogsetA}k & \cellcolor{cogsetA}u & \cellcolor{cogsetA}m & \cellcolor{cogsetA}- & \cellcolor{cogsetA}- & \cellcolor{cogsetA}- & \cellcolor{cogsetA}I \\
        Udihe    & 16\_lousen-37 & \cellcolor{cogsetA}- & \cellcolor{cogsetA}k & \cellcolor{cogsetA}u & \cellcolor{cogsetA}m & \cellcolor{cogsetA}u & \cellcolor{cogsetA}x & \cellcolor{cogsetA}- & \cellcolor{cogsetA}I \\
        Nanai    & 16\_lousen-38 & \cellcolor{cogsetB}- & \cellcolor{cogsetB}t & \cellcolor{cogsetB}- & \cellcolor{cogsetB}- & \cellcolor{cogsetB}i & \cellcolor{cogsetB}k & \cellcolor{cogsetB}t & \cellcolor{cogsetB}I \\
        Orok     & 16\_lousen-38 & \cellcolor{cogsetB}- & \cellcolor{cogsetB}t & \cellcolor{cogsetB}- & \cellcolor{cogsetB}- & \cellcolor{cogsetB}i & \cellcolor{cogsetB}k & \cellcolor{cogsetB}t & \cellcolor{cogsetB}I \\
        Ulch     & 16\_lousen-38 & \cellcolor{cogsetB}- & \cellcolor{cogsetB}t & \cellcolor{cogsetB}- & \cellcolor{cogsetB}- & \cellcolor{cogsetB}i & \cellcolor{cogsetB}q & \cellcolor{cogsetB}t & \cellcolor{cogsetB}I \\\bottomrule
    \end{tabular}~~~~\begin{tabular}{lcccc}
        \toprule
        \textbf{Language} & \textbf{sound class} & \textbf{k} & \textbf{q} & \textbf{x} \\\midrule
        Even    & 1 & 1 & 0 & 0 \\
        Kilen   & 1 & 1 & 0 & 0 \\
        Negidal & 1 & 1 & 0 & 0 \\
        Oroch   & 0 & 0 & 0 & 0 \\
        Udihe   & 1 & 0 & 0 & 1 \\
        Nanai   & 1 & 1 & 0 & 0 \\
        Orok    & 1 & 1 & 0 & 0 \\
        Ulch    & 1 & 0 & 1 & 0 \\\bottomrule
    \end{tabular}
    \caption{Example of a multiple sequence alignment. Alignment cells are shaded to indicate different cognate sets. (left) Binarized version of column 6. (right)}
    \label{tab:1}
\end{table*}

As can be seen from this example, the MSA contains information about cognacy, but also about sound correspondences. For example, a \texttt{t} in column 2 is a proxy for the cognate class \texttt{16\_lousen-38}, while \texttt{k} and \texttt{h} in the same column correspond to the cognate class \texttt{16\_lousen-37}. In column 6, the cognate class \texttt{16\_lousen-38} is split into two sound classes, \texttt{k} and \texttt{q}, reflecting a sound change; likewise, Udihe has \texttt{x} where the other reflexes of \texttt{16\_lousen-37} have \texttt{k}. The presence of a sound class, as opposed to a gap, can also serve as a proxy of a cognate class: in column 7, e.g., \texttt{t} is present in exactly the reflexes of \texttt{16\_lousen-38}. Put differently, binary characters corresponding to a gap are flipped by switching 0s and 1s.

In a fifth step, the MSA was converted to a binary matrix. Two binarization methods were used simultaneously. For a given column in an MSA, a character was created for the presence of a sound class. For column 6 in Table \ref{tab:1}, e.g., this character has value $1$ for all languages except Oroch. Additionally, for each sound class type in a column, a different character was created. For column 6, there are three such characters, one for \texttt{k}, one for \texttt{q}, and one for \texttt{x}. The first has value $1$ for Even, Kilen, Negidal, Nanai and Orok and $0$ otherwise; the second has value $1$ only for Ulch, and the third only for Udihe. Languages for which the data do not contain a reflex for a given concept are coded as missing for all relevant characters. If a language has multiple reflexes for a given concept, the maximum value is chosen. 

Applying this workflow to all concepts and concatenating the resulting matrices yields the final character matrix with 974 rows and 62,512 columns.

As mentioned above, this workflow builds on the method by \citet{akavarapu-bhattacharya-2024-likelihood}, but differs in various aspects. The mentioned work (1) uses Dolgopolsky sound classes instead of ASJP, (2) finds the MSA using CLUSTALW2 \citep{larkin2007clustal} instead of T-Coffee, and (3) omits the binarization steps, working with a multi-state model of evolution for phylogenetic inference.

In the tables and figures this method is referred to as MSA.

\subsubsection{Phylogenetic inference}

We performed phylogenetic inference using \emph{raxml-ng} \citep{kozlov2019raxml}, which implements maximum-likelihood estimation. The BIN+G model (binary substitution model with gamma-distributed rates) was used for all analyses. This means that gain rates and loss rates can be different, and that the mutation rates of the different characters can differ but are drawn from the same gamma distribution. The parameters of this distribution are estimated from the data.

Version 2.0.2 of \emph{raxml-ng} was used in its adaptive mode, where the number of starting trees (both random and maximum-parsimony trees) is chosen automatically based on the predicted difficulty of the character matrix. All tree searches were run with a fixed random seed. The tree with the highest likelihood was chosen as the final result.

\begin{table*}[t!]
    \centering
    \begin{tabular}{lS[table-format=1.3] r S[table-format=2.3]}
        \toprule
        \textbf{Method} & \textbf{GQD (Glottolog)} & \textbf{AIC (Grambank)}& \textbf{difficulty}\\
        \midrule
        Cognate classes         & 0.194          & 109,584 & 0.38 \\
        PMI      & 0.092     & 109,701 & 0.47 \\
        MSA                     & \textbf{0.036} & \textbf{109,141} & \textbf{0.25} \\\bottomrule
    \end{tabular}
    \caption{Evaluation of the full dataset. GQD = Generalized Quartet Distance (lower is better; ranges from 0 for perfect fit to $0.67$ for chance level); AIC = Akaike Information Criterion for typological model fit (lower is better; absolute values are not interpretable in isolation but differences are meaningful); difficulty = phylogenetic difficulty estimated with Pythia (lower is better; ranges from 0 for strong phylogenetic signal to 1 for absent signal).}

    \label{tab:2}
\end{table*}

\begin{table*}[t!]
    \centering
    \begin{tabular}{lS[table-format=1.3] S[table-format=1.3] S[table-format=4.0] S[table-format=3.0] S[table-format=1.3] S[table-format=1.3]}
        \toprule
        \textbf{Method} & \textbf{$\mu$ GQD} & \textbf{$\sigma$ GQD} &\textbf{$\mu$ AIC}& \textbf{$\sigma$ AIC} & \textbf{$\mu$ difficulty}&\textbf{$\sigma$ difficulty}\\\midrule
        Cognate classes & 0.269 & 0.082 & 167& 99 &0.302 &0.049\\
        PMI & 0.144 & 0.071 &-34 & 72 & 0.148&0.049\\
        MSA & \textbf{0.038} & 0.013 &\textbf{-133}&58 &\textbf{0.114} &0.041\\\bottomrule
    \end{tabular}
    \caption{Evaluation of the 100 random samples ($\mu$: sample mean; $\sigma$: sample standard deviation).}
    \label{tab:3}
\end{table*}

\begin{table*}[t!]
    \centering
    \begin{tabular}{l S[table-format=1.3] S[table-format=1.3] S[table-format=2.2] S[table-format=2.2] S[table-format=1.3] S[table-format=1.3]}
        \toprule
        \textbf{Method} & \textbf{$\mu$ GQD} & \textbf{$\sigma$ GQD} &\textbf{$\mu$ AIC}& \textbf{$\sigma$ AIC} & \textbf{$\mu$ difficulty}&\textbf{$\sigma$ difficulty}\\\midrule
        Cognate classes & 0.224          & 0.144 & 6.39& 53.22 &0.322 &0.205\\
        PMI             & 0.217          & 0.118 &-2.02 & 22.99 & 0.234&0.175\\
        MSA             & \textbf{0.190} & 0.114 &\textbf{-4.37}&32.09 &\textbf{0.139} &0.116\\\bottomrule
    \end{tabular}
    \caption{Evaluation of the 15 largest language families ($\mu$: sample mean; $\sigma$: sample standard deviation).}
    \label{tab:4}
\end{table*}
\begin{figure*}[t!]
    \centering
    \includegraphics[width=.45\linewidth, height=0.85\textheight, keepaspectratio]{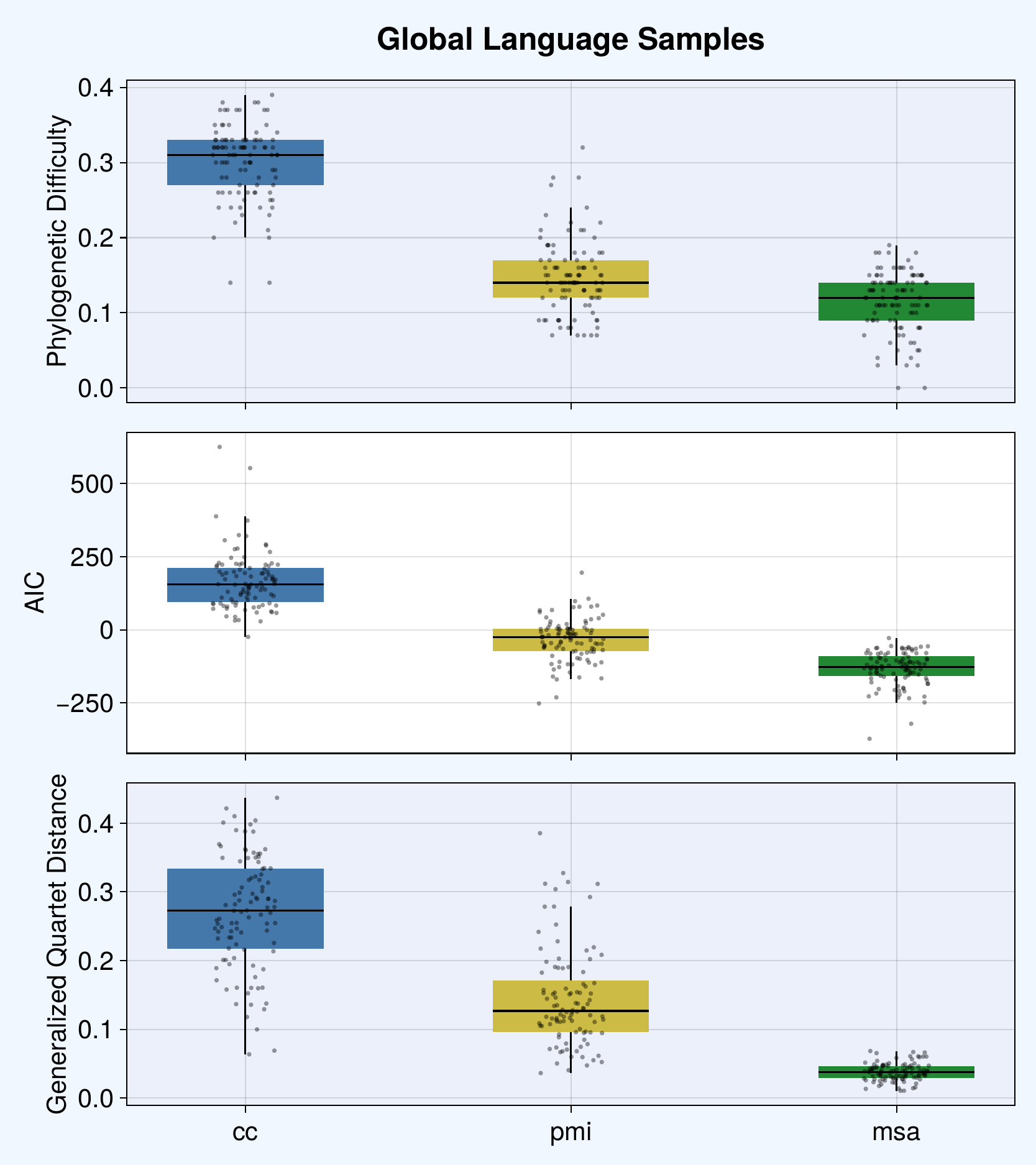}\hspace{\fill}\includegraphics[width=.45\linewidth, height=0.85\textheight, keepaspectratio]{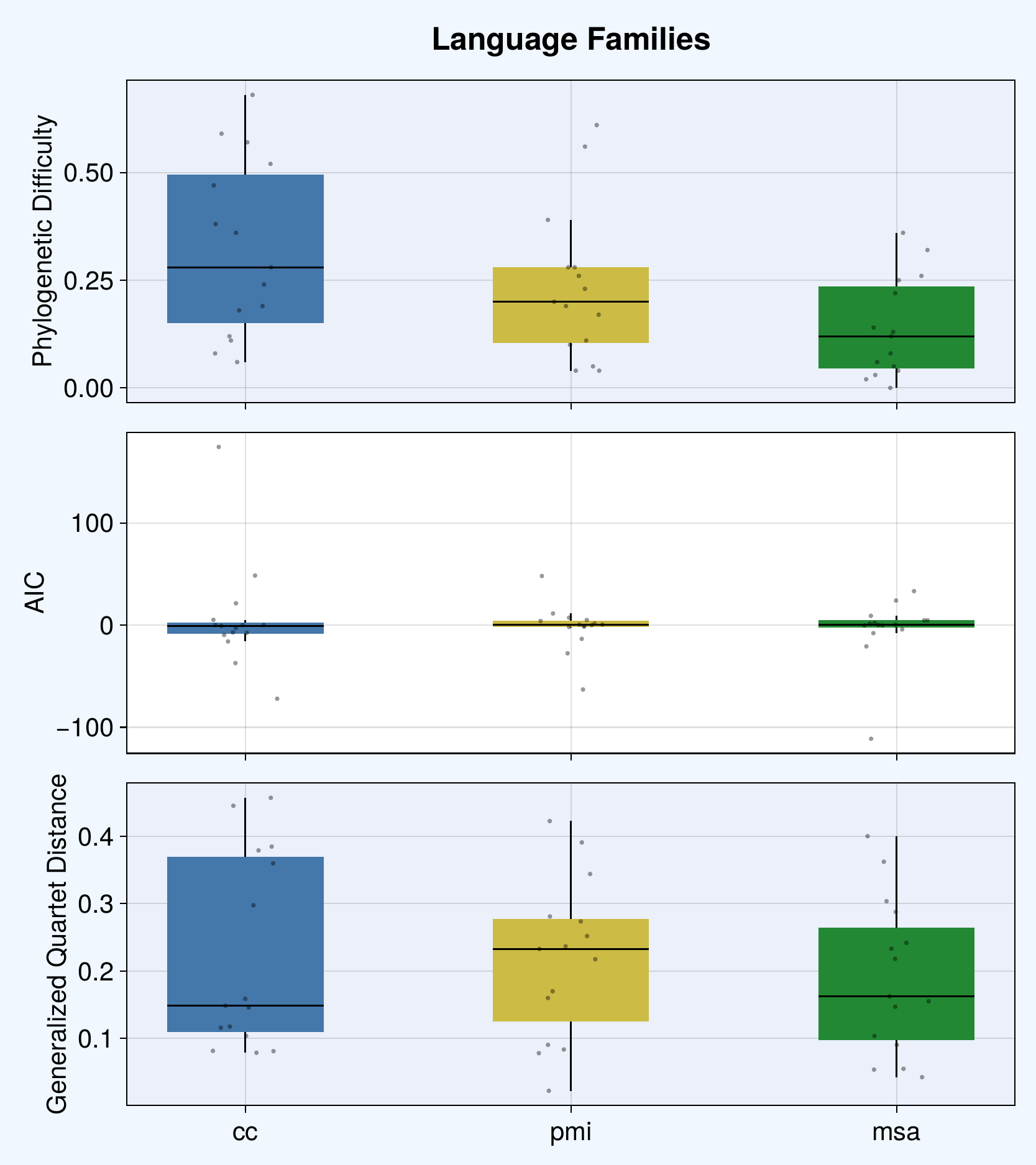}
    \caption{\textbf{Left panel:} Comparison of methods across three evaluation metrics for the 100 random samples. The boxplots show distribution per method, while the overlaid points represent individual samples. \textbf{Right panel:} Comparison of methods across three evaluation metrics for the 15 largest language families. The boxplots show distribution per method, while the overlaid points represent individual samples.}
    \label{fig:1}
\end{figure*}

\subsubsection{Evaluation}

Evaluation was conducted on three types of datasets:

\begin{itemize}
    \item the full dataset of 974 languages,
    \item 100 samples of 100 languages each, which are drawn at random without replacement from the full dataset, and
    \item a collection of 15 language families, each with at least 10 languages.
\end{itemize}

For each of these groups of datasets, the following evaluations were performed:

\paragraph{Comparison with Glottolog}

The Glottolog classification of the languages in a dataset can be represented as a phylogenetic tree with polytomies, i.e., with nodes containing more than two daughters. This Glottolog tree serves as gold standard. To assess the degree of agreement between the gold standard and the inferred phylogenies, the \emph{generalized quartet distance} (GQD) was deployed, as first proposed by \citet{pompeieatl11}. This distance is defined as the fraction of quartets (i.e., sets of four languages) that are (a) resolved in both trees, and (b) resolved differently in the two trees. The GQD ranges from 0 (perfect agreement) to $0.67$ (chance level). The GQD was computed using the software \emph{QDist}, which can be obtained from \url{https://birc.au.dk/software/qdist/}.

\paragraph{Fit with Grambank}

The hypothesis is assumed that the values of the Grambank features evolve along a phylogeny in the same way as the lexical characters described earlier in this section. The degree of fit of the inferred phylogenies with the Grambank features was assessed by (1) using the inferred phylogeny and estimating the branch lengths, mutation rates and rate heterogeneity via Maximum Likelihood, and (2) computing the \emph{Akaike Information Criterion} (AIC). A lower AIC value indicates a better fit.

ML inference and AIC computation were also performed with \emph{raxml-ng}.

For the groups of random samples and of language families, AIC values were normalized to mean 0 to facilitate comparison.

\paragraph{Phylogenetic difficulty}

The strength of the phylogenetic signal in the data was assessed using the \emph{Pythia} difficulty score \citep{haag2022easy}. The authors define a measure of signal strength that uses 100 maximum likelihood tree searches and quantifies the degree of agreement between the inferred trees. Pythia is a machine learning algorithm that predicts this difficulty from various properties of the character matrix, such as entropy and sites-over-taxa ratio, and maximum-parsimony tree inference, with high precision and comparatively low computational cost. The measure ranges from 0 (little difficulty, i.e., strongest signal) to 1 (very difficult, i.e., no signal). The difficulty scores reported here were computed with the Pythia predictor built into \emph{raxml-ng} 2.0.2, using a fixed random seed. Predicted difficulties are not comparable across versions of the predictor.

\section{Results}

The evaluation results for the entire dataset are shown in Table \ref{tab:2}. Table \ref{tab:3} shows the aggregated results for the 100 random samples. They are visualized in the left panel of Figure \ref{fig:1}.

The aggregated evaluation results for the 15 largest language families are shown in Table \ref{tab:4}. They are visualized in the right panel of Figure \ref{fig:1}. The results for the individual families are given in Table \ref{tab:5}.

\begin{table*}
    \centering
        \begin{tabular}{llrrr}
            \toprule
            \rowcolor{white} \textbf{Family} & \textbf{Metric} & \textbf{cc} & \textbf{pmi} & \textbf{msa} \\\midrule
            \rowcolor{white} Afro-Asiatic & GQD & 0.385 & 0.170 & \textbf{0.162} \\
            \rowcolor{white}  & PhyDiff & 0.240 & 0.110 & \textbf{0.040} \\
            \rowcolor{white}  & AIC & 48.478 & \textbf{-27.630} & -20.847 \\
            \rowcolor{gray!20} Arawakan & GQD & 0.116 & 0.281 & \textbf{0.043} \\
            \rowcolor{gray!20}  & PhyDiff & 0.180 & 0.200 & \textbf{0.020} \\
            \rowcolor{gray!20}  & AIC & \textbf{-16.035} & 11.354 & 4.681 \\
            \rowcolor{white} Atlantic-Congo & GQD & \textbf{0.146} & 0.218 & 0.288 \\
            \rowcolor{white}  & PhyDiff & 0.590 & 0.610 & \textbf{0.250} \\
            \rowcolor{white}  & AIC & \textbf{-37.200} & 4.006 & 33.195 \\
            \rowcolor{gray!20} Austroasiatic & GQD & 0.079 & 0.078 & \textbf{0.055} \\
            \rowcolor{gray!20}  & PhyDiff & \textbf{0.120} & 0.230 & 0.140 \\
            \rowcolor{gray!20}  & AIC & 21.362 & \textbf{-13.488} & -7.874 \\
            \rowcolor{white} Austronesian & GQD & \textbf{0.104} & 0.160 & 0.242 \\
            \rowcolor{white}  & PhyDiff & 0.680 & 0.560 & \textbf{0.360} \\
            \rowcolor{white}  & AIC & \textbf{-72.059} & 48.052 & 24.007 \\
            \rowcolor{gray!20} Cariban & GQD & 0.360 & 0.422 & \textbf{0.304} \\
            \rowcolor{gray!20}  & PhyDiff & \textbf{0.280} & \textbf{0.280} & 0.320 \\
            \rowcolor{gray!20}  & AIC & -0.104 & 0.527 & \textbf{-0.423} \\
            \rowcolor{white} Chibchan & GQD & \textbf{0.159} & 0.391 & 0.362 \\
            \rowcolor{white}  & PhyDiff & 0.380 & 0.390 & \textbf{0.060} \\
            \rowcolor{white}  & AIC & \textbf{-7.178} & -1.860 & 9.038 \\
            \rowcolor{gray!20} Dravidian & GQD & 0.298 & 0.237 & \textbf{0.218} \\
            \rowcolor{gray!20}  & PhyDiff & 0.360 & 0.170 & \textbf{0.050} \\
            \rowcolor{gray!20}  & AIC & -0.076 & 0.470 & \textbf{-0.394} \\
            \rowcolor{white} Indo-European & GQD & 0.082 & \textbf{0.022} & 0.054 \\
            \rowcolor{white}  & PhyDiff & 0.190 & \textbf{0.100} & 0.120 \\
            \rowcolor{white}  & AIC & \textbf{-7.536} & 7.253 & 0.283 \\
            \rowcolor{gray!20} Pama-Nyungan & GQD & \textbf{0.149} & 0.344 & 0.155 \\
            \rowcolor{gray!20}  & PhyDiff & 0.570 & \textbf{0.260} & \textbf{0.260} \\
            \rowcolor{gray!20}  & AIC & \textbf{-9.669} & 4.994 & 4.675 \\
            \rowcolor{white} Sino-Tibetan & GQD & 0.457 & 0.233 & \textbf{0.147} \\
            \rowcolor{white}  & PhyDiff & 0.520 & 0.190 & \textbf{0.130} \\
            \rowcolor{white}  & AIC & 174.326 & -63.064 & \textbf{-111.262} \\
            \rowcolor{gray!20} Tucanoan & GQD & 0.379 & \textbf{0.274} & 0.400 \\
            \rowcolor{gray!20}  & PhyDiff & 0.080 & 0.040 & \textbf{0.030} \\
            \rowcolor{gray!20}  & AIC & \textbf{-2.820} & 1.480 & 1.339 \\
            \rowcolor{white} Tupian & GQD & 0.445 & 0.252 & \textbf{0.233} \\
            \rowcolor{white}  & PhyDiff & 0.470 & 0.280 & \textbf{0.220} \\
            \rowcolor{white}  & AIC & -0.767 & \textbf{-1.428} & 2.196 \\
            \rowcolor{gray!20} Turkic & GQD & \textbf{0.081} & 0.091 & 0.091 \\
            \rowcolor{gray!20}  & PhyDiff & 0.060 & \textbf{0.050} & 0.080 \\
            \rowcolor{gray!20}  & AIC & -0.011 & \textbf{-0.045} & 0.056 \\
            \rowcolor{white} Uto-Aztecan & GQD & 0.118 & \textbf{0.084} & 0.104 \\
            \rowcolor{white}  & PhyDiff & 0.110 & 0.040 & \textbf{0.000} \\
            \rowcolor{white}  & AIC & 5.122 & -0.916 & \textbf{-4.207} \\\bottomrule
        \end{tabular}
            \caption{Evaluation of the 15 largest language families. The best value for each family is highlighted in bold.}  
            \label{tab:5}    
\end{table*}

When focusing on phylogenetic inference at the level of individual families, we find a considerable variation between families. This applies both to the numerical evaluation results and the relative ranking of the three methods considered here. The MSA method tends to produce the lowest phylogenetic difficulty, while there is no discernible trend regarding the fit to Glottolog and to Grambank.

This picture changes considerably when we focus on datasets covering languages from many different families. Here, the MSA method consistently outperforms the other two methods. This is particularly evident in the comparison with Glottolog, where the MSA method yields the lowest GQD values. The MSA method also leads to the lowest AIC values, indicating a better fit to the Grambank typological features. The phylogenetic difficulty is also lowest for the MSA method.

\section{Discussion}

These findings suggest that the MSA method is a promising alternative to traditional cognate-based methods. It is competitive with the more labor-intensive method based on manual cognate annotations, as well as the method using automatically detected cognate classifications, when considering individual language families. For global datasets, the MSA method clearly outperforms the other two methods. This is particularly evident in the comparison with Glottolog, where the MSA method yields the lowest GQD values. The MSA method also tends to produce the lowest AIC values, indicating a better fit to the Grambank typological features. The phylogenetic difficulty is also lowest for the MSA method.

\section*{Limitations}

The two evaluation methods that quantify the fit of the inferred trees to empirical data only assess the quality of the inferred tree \textbf{topologies}. Future work will need to address the question how well the inferred branch lengths and divergence dates correspond to the true values. This is a challenging task, as the true values are unknown. It is expected that the usefulness for downstream tasks is a suitable proxy.

\section*{Data and Code Availability}

The code used in this study is available at \url{https://codeberg.org/profgerhard/sigtyp2025_code/}. The complete data and code, including all primary and intermediate datasets, are archived in FDAT, the research data repository of the University of Tübingen, at \url{https://doi.org/10.57754/FDAT.8n2vd-1bq37}.

\section*{Acknowledgments}

This research was supported by the DFG Centre for Advanced Studies in the Humanities Words, Bones, Genes, Tools (DFG-KFG 2237) and by the European Research Council (ERC) under the European Union's Horizon 2020 research and innovation programme (Grant agreement 834050).

Claude Code (Anthropic) assisted with diagnosing and fixing the implementation defects described in the title footnote, re-running the analysis pipeline, and preparing this corrected version.

\bibliography{references}

\begin{thebibliography}{25}
\providecommand{\natexlab}[1]{#1}

\bibitem[{Akavarapu and
  Bhattacharya(2024)}]{akavarapu-bhattacharya-2024-likelihood}
V.S.D.S.Mahesh Akavarapu and Arnab Bhattacharya. 2024.
\newblock \href {https://doi.org/10.18653/v1/2024.naacl-long.141} {A likelihood
  ratio test of genetic relationship among languages}.
\newblock In \emph{Proceedings of the 2024 Conference of the North American
  Chapter of the Association for Computational Linguistics: Human Language
  Technologies (Volume 1: Long Papers)}, pages 2559--2570, Mexico City, Mexico.
  Association for Computational Linguistics.

\bibitem[{Bentz et~al.(2018)Bentz, Dediu, Verkerk, and
  J{\"a}ger}]{bentz2018evolution}
Christian Bentz, Dan Dediu, Annemarie Verkerk, and Gerhard J{\"a}ger. 2018.
\newblock The evolution of language families is shaped by the environment
  beyond neutral drift.
\newblock \emph{Nature Human Behaviour}, 2(11):816--821.

\bibitem[{Bouckaert et~al.(2022)Bouckaert, Redding, Sheehan, Kyritsis, Gray,
  Jones, and Atkinson}]{bouckaert2022global}
Remco Bouckaert, David Redding, Oliver Sheehan, Thanos Kyritsis, Russel Gray,
  Kate~E Jones, and Quentin Atkinson. 2022.
\newblock Global language diversification is linked to socio-ecology and threat
  status.
\newblock \emph{SocArXiv}.

\bibitem[{Durbin et~al.(1998)Durbin, Eddy, Krogh, and
  Mitchison}]{durbin1998biological}
Richard Durbin, Sean~R Eddy, Anders Krogh, and Graeme Mitchison. 1998.
\newblock \emph{Biological sequence analysis: probabilistic models of proteins
  and nucleic acids}.
\newblock Cambridge university press.

\bibitem[{Gascuel(1997)}]{gascuel1997bionj}
Olivier Gascuel. 1997.
\newblock {BIONJ}: an improved version of the {NJ} algorithm based on a simple
  model of sequence data.
\newblock \emph{Molecular Biology and Evolution}, 14(7):685--695.

\bibitem[{Gray and Atkinson(2003)}]{gray2003language}
Russell~D Gray and Quentin~D Atkinson. 2003.
\newblock Language-tree divergence times support the anatolian theory of
  indo-european origin.
\newblock \emph{Nature}, 426(6965):435--439.

\bibitem[{Haag et~al.(2022)Haag, H{\"o}hler, Bettisworth, and
  Stamatakis}]{haag2022easy}
Julia Haag, Dimitri H{\"o}hler, Ben Bettisworth, and Alexandros Stamatakis.
  2022.
\newblock From easy to hopeless—predicting the difficulty of phylogenetic
  analyses.
\newblock \emph{Molecular biology and evolution}, 39(12):msac254.

\bibitem[{Hammarstr{\"o}m et~al.(2024)Hammarstr{\"o}m, Forkel, Haspelmath, and
  Bank}]{hammarstrom2024glottolog}
Harald Hammarstr{\"o}m, Robert Forkel, Martin Haspelmath, and Sebastian Bank.
  2024.
\newblock Glottolog 5.1.
\newblock \emph{Leipzig: Max Planck Institute for Evolutionary
  Anthropology.(Available online atglottolog. org, Accessed on2024-11-29.)},
  10.

\bibitem[{Heggarty et~al.(2023)Heggarty, Anderson, Scarborough, King,
  Bouckaert, Jocz, K\"{u}mmel, J\"{u}gel, Irslinger, Pooth, Liljegren, Strand,
  Haig, Mac{\'{a}}k, Kim, Anonby, Pronk, Belyaev, Dewey-Findell, Boutilier,
  Freiberg, Tegethoff, Serangeli, Liosis, Stro{\'{n}}ski, Schulte, Gupta, Haak,
  Krause, Atkinson, Greenhill, K\"{u}hnert, and Gray}]{Heggarty2023}
Paul Heggarty, Cormac Anderson, Matthew Scarborough, Benedict King, Remco
  Bouckaert, Lechos{\l}aw Jocz, Martin~Joachim K\"{u}mmel, Thomas J\"{u}gel,
  Britta Irslinger, Roland Pooth, Henrik Liljegren, Richard~F. Strand, Geoffrey
  Haig, Martin Mac{\'{a}}k, Ronald~I. Kim, Erik Anonby, Tijmen Pronk, Oleg
  Belyaev, Tonya~Kim Dewey-Findell, and 14 others. 2023.
\newblock \href {https://doi.org/10.1126/science.abg0818} {Language trees with
  sampled ancestors support a hybrid model for the origin of {Indo-European}
  languages}.
\newblock \emph{Science}, 381(6656).

\bibitem[{J{\"a}ger(2015)}]{jager2015support}
Gerhard J{\"a}ger. 2015.
\newblock Support for linguistic macrofamilies from weighted sequence
  alignment.
\newblock \emph{Proceedings of the National Academy of Sciences},
  112(41):12752--12757.

\bibitem[{J{\"a}ger(2018)}]{jager2018global}
Gerhard J{\"a}ger. 2018.
\newblock Global-scale phylogenetic linguistic inference from lexical
  resources.
\newblock \emph{Scientific Data}, 5(1):1--16.

\bibitem[{Kolipakam et~al.(2018)Kolipakam, Jordan, Dunn, Greenhill, Bouckaert,
  Gray, and Verkerk}]{Kolipakam2018}
Vishnupriya Kolipakam, Fiona~M. Jordan, Michael Dunn, Simon~J. Greenhill, Remco
  Bouckaert, Russell~D. Gray, and Annemarie Verkerk. 2018.
\newblock A {Bayesian} phylogenetic study of the {Dravidian} language family.
\newblock \emph{Royal Society Open Science}, 5(171504):1--17.

\bibitem[{Kozlov et~al.(2019)Kozlov, Darriba, Flouri, Morel, and
  Stamatakis}]{kozlov2019raxml}
Alexey~M Kozlov, Diego Darriba, Tom{\'a}{\v{s}} Flouri, Benoit Morel, and
  Alexandros Stamatakis. 2019.
\newblock Raxml-ng: a fast, scalable and user-friendly tool for maximum
  likelihood phylogenetic inference.
\newblock \emph{Bioinformatics}, 35(21):4453--4455.

\bibitem[{Larkin et~al.(2007)Larkin, Blackshields, Brown, Chenna, McGettigan,
  McWilliam, Valentin, Wallace, Wilm, Lopez et~al.}]{larkin2007clustal}
Mark~A Larkin, Gordon Blackshields, Nigel~P Brown, R~Chenna, Paul~A McGettigan,
  Hamish McWilliam, Franck Valentin, Iain~M Wallace, Andreas Wilm, Rodrigo
  Lopez, and 1 others. 2007.
\newblock Clustal w and clustal x version 2.0.
\newblock \emph{bioinformatics}, 23(21):2947--2948.

\bibitem[{List and Forkel(2024)}]{LingPy2024}
Johann-Mattis List and Robert Forkel. 2024.
\newblock \href
  {https://doi.org/https://zenodo.org/badge/latestdoi/5137/lingpy/lingpy}
  {Lingpy. a python library for historical linguistics}.
\newblock With contributions by Simon Greenhill, Tiago Tresoldi, Christoph
  Rzymski, Gereon Kaiping, Steven Moran, Peter Bouda, Johannes Dellert, Taraka
  Rama, Frank Nagel, Patrick Elmer, Arne Rubehn.

\bibitem[{List et~al.(2023)List, Forkel, Greenhill, Rzymski, Englisch, and
  Gray}]{List2023}
Johann-Mattis List, Robert Forkel, Simon~J. Greenhill, Christoph Rzymski,
  Johannes Englisch, and Russell~D. Gray. 2023.
\newblock \href {https://doi.org/10.5281/zenodo.7836668} {Lexibank analysed}.
\newblock \emph{Scientific Data}, 9(316):1--31.
\newblock Data set.

\bibitem[{Notredame et~al.(2000)Notredame, Higgins, and
  Heringa}]{notredame2000t}
C{\'e}dric Notredame, Desmond~G Higgins, and Jaap Heringa. 2000.
\newblock T-coffee: A novel method for fast and accurate multiple sequence
  alignment.
\newblock \emph{Journal of molecular biology}, 302(1):205--217.

\bibitem[{Oskolskaya et~al.(2021)Oskolskaya, Koile, and
  Robbeets}]{Oskolskaya2021}
S.~Oskolskaya, E.~Koile, and M.~Robbeets. 2021.
\newblock \href {https://doi.org/10.1075/dia.20010.osk} {A {Bayesian} approach
  to the classification of {Tungusic} languages}.
\newblock \emph{Diachronica}, 39(1):128--158.

\bibitem[{Pompei et~al.(2011)Pompei, Loreto, and Tria}]{pompeieatl11}
Simone Pompei, Vittorio Loreto, and Francesca Tria. 2011.
\newblock On the accuracy of language trees.
\newblock \emph{PLoS One}, 6(6):e20109.

\bibitem[{Raghavan et~al.(2007)Raghavan, Albert, and Kumara}]{raghavan2007near}
Usha~Nandini Raghavan, R{\'e}ka Albert, and Soundar Kumara. 2007.
\newblock Near linear time algorithm to detect community structures in
  large-scale networks.
\newblock \emph{Physical Review E—Statistical, Nonlinear, and Soft Matter
  Physics}, 76(3):036106.

\bibitem[{Rama et~al.(2018)Rama, List, Wahle, and
  J{\"a}ger}]{rama2018automatic}
Taraka Rama, Johann-Mattis List, Johannes Wahle, and Gerhard J{\"a}ger. 2018.
\newblock Are automatic methods for cognate detection good enough for
  phylogenetic reconstruction in historical linguistics?
\newblock In \emph{Proceedings of the 2018 Conference of the North American
  Chapter of the Association for Computational Linguistics: Human Language
  Technologies, Volume 2 (Short Papers)}, pages 393--400.

\bibitem[{Ringe et~al.(2002)Ringe, Warnow, and Taylor}]{Ringe2002}
Donald Ringe, Tandy Warnow, and Ann Taylor. 2002.
\newblock {I}ndo-{E}uropean and computational cladistics.
\newblock \emph{Transactions of the Philological Society}, 100(1):59--129.

\bibitem[{Sagart et~al.(2019)Sagart, Jacques, Lai, Ryder, Thouzeau, Greenhill,
  and List}]{Sagart2019}
Laurent Sagart, Guillaume Jacques, Yunfan Lai, Robin Ryder, Valentin Thouzeau,
  Simon~J. Greenhill, and Johann-Mattis List. 2019.
\newblock \href {https://doi.org/10.1073/pnas.1817972116} {Dated language
  phylogenies shed light on the ancestry of {Sino-Tibetan}}.
\newblock \emph{Proceedings of the National Academy of Science of the United
  States of America}, 116:10317--10322.

\bibitem[{Skirg{\aa}rd et~al.(2023)Skirg{\aa}rd, Haynie, Blasi,
  Hammarstr{\"o}m, Collins, Latarche, Lesage, Weber, Witzlack-Makarevich,
  Passmore, Chira, Maurits, Dinnage, Dunn, Reesink, Singer, Bowern, Epps, Hill,
  Vesakoski, Robbeets, Abbas, Auer, Bakker, Barbos, Borges, Danielsen,
  Dorenbusch, Dorn, Elliott, Falcone, Fischer, Ghanggo~Ate, Gibson, G{\"o}bel,
  Goodall, Gruner, Harvey, Hayes, Heer, Herrera~Miranda, H{\"u}bler,
  Huntington-Rainey, Ivani, Johns, Just, Kashima, Kipf, Klingenberg, K{\"o}nig,
  Koti, Kowalik, Krasnoukhova, Lindvall, Lorenzen, Lutzenberger, Martins,
  Mata~German, van~der Meer, Montoya~Samam{\'e}, M{\"u}ller,
  Murado$\breve{g}$lu, Neely, Nickel, Norvik, Oluoch, Peacock, Pearey, Peck,
  Petit, Pieper, Poblete, Prestipino, Raabe, Raja, Reimringer, Rey, Rizaew,
  Ruppert, Salmon, Sammet, Schembri, Schlabbach, Schmidt, Skilton, Smith,
  de~Sousa, Sverredal, Valle, Vera, Vo{\ss}, Witte, Wu, Yam, Ye~葉婧婷,
  Yong, Yuditha, Zariquiey, Forkel, Evans, Levinson, Haspelmath, Greenhill,
  Atkinson, and Gray}]{grambank_release}
Hedvig Skirg{\aa}rd, Hannah~J. Haynie, Dami{\'a}n~E. Blasi, Harald
  Hammarstr{\"o}m, Jeremy Collins, Jay~J. Latarche, Jakob Lesage, Tobias Weber,
  Alena Witzlack-Makarevich, Sam Passmore, Angela Chira, Luke Maurits, Russell
  Dinnage, Michael Dunn, Ger Reesink, Ruth Singer, Claire Bowern, Patience
  Epps, Jane Hill, and 86 others. 2023.
\newblock \href {https://doi.org/10.1126/sciadv.adg6175} {Grambank reveals the
  importance of genealogical constraints on linguistic diversity and highlights
  the impact of language loss}.
\newblock \emph{Science Advances}, 9.

\bibitem[{Tria et~al.(2017)Tria, Landan, and Dagan}]{tria2017phylogenetic}
Fernando Domingues~K{\"u}mmel Tria, Giddy Landan, and Tal Dagan. 2017.
\newblock Phylogenetic rooting using minimal ancestor deviation.
\newblock \emph{Nature Ecology \& Evolution}, 1(1):0193.

\end{thebibliography}
  
%   \appendix
  
%   \section{Example Appendix}
%   \label{sec:appendix}
  
%   This is an appendix.
  
  \end{document}